\documentclass[11pt]{article}

\usepackage[preprint]{acl}

\usepackage{times}
\usepackage{latexsym}

\usepackage[T1]{fontenc}

\usepackage[utf8]{inputenc}

\usepackage{microtype}

\usepackage{inconsolata}

\usepackage{graphicx}

\usepackage{amsmath}
\usepackage{mathtools}
\usepackage{amssymb}
\usepackage{amsfonts}
\usepackage{bm}
\usepackage{bbm}

\usepackage{booktabs}
\usepackage{tabularx}
\usepackage{nicematrix}
\usepackage{multirow}
\usepackage{colortbl}
\definecolor{lightgreen}{rgb}{0.9, 1.0, 0.9}
\usepackage[subrefformat=parens]{subcaption}
\usepackage{pifont}
\usepackage{accsupp}

\DeclareUnicodeCharacter{FE0F}{%
  \BeginAccSupp{method=hex,unicode,ActualText=FE0F}%
    \null
  \EndAccSupp{}%
}

\usepackage[linesnumbered,ruled,lined,noend]{algorithm2e}
\SetKwInOut{Given}{Given\hspace{-0em}}
\SetKwInOut{Input}{Input\hspace{0em}}
\SetKwInOut{Output}{Output\hspace{0em}}
\SetKwFor{Foreach}{for each}{do}{}
\SetKw{Continue}{continue}
\SetKw{Break}{break}
\DontPrintSemicolon
\SetInd{0.5em}{0.5em}

\usepackage{cleveref}
\crefname{equation}{Equation}{Equation}
\crefname{figure}{Figure}{Figure}
\crefname{table}{Table}{Table}
\crefname{section}{Section}{Section}
\crefname{subsection}{Section}{Section}
\crefname{appendix}{Appendix}{Appendix}
\crefname{algorithm}{Algorithm}{Algorithm}
\crefname{line}{Line}{Line}

\DeclareMathOperator*{\argmax}{argmax}
\DeclareMathOperator*{\argmin}{argmin}

\newcommand{\styMetric}[1]{\textsc{#1}}

\newcommand{\metricComet}{\styMetric{Comet}}

\newcommand{\metricClipscore}{\styMetric{ClipScore}}

\newcommand{\trademark}{${}^{\text{\texttrademark}}$}

\newcommand{\styVector}[1]{\mathbf{#1}}
\newcommand{\stySet}[1]{\mathcal{#1}}

\newcommand{\R}{\mathbb{R}}
\newcommand{\N}{\mathbb{N}}

\newcommand{\inputText}{\styVector{w}}
\newcommand{\inputImage}{\styVector{I}}
\newcommand{\funcLoss}{\mathcal{L}}
\newcommand{\funcObj}{\mathcal{J}}
\newcommand{\score}{S}

\newcommand{\sizeHeight}{H}
\newcommand{\sizeWidth}{W}
\newcommand{\sizeChannel}{C}
\newcommand{\sizeDim}{D}
\newcommand{\constScale}{M}
\newcommand{\sizeBeam}{k}

\newcommand{\textHub}{\styVector{w}}

\newcommand{\embedding}[1]{\styVector{v}_{#1}}

\newcommand{\funcEncoder}{f}
\newcommand{\paramEncoder}{\theta}
\newcommand{\funcClipScore}{s}
\newcommand{\funcTopk}{\text{Top-}k}
\newcommand{\paramInverter}{\phi}

\newcommand{\spaceText}{\vocab^\ast}
\newcommand{\spaceImage}{\stySet{I}}
\newcommand{\setHypotheses}{\stySet{H}}
\newcommand{\vocab}{\stySet{V}}
\newcommand{\setData}{\stySet{D}}
\newcommand{\setDataImage}{\setData_{\spaceImage}}
\newcommand{\setDataText}{\setData_{\spaceText}}

\newcommand{\setBeam}{\stySet{B}}
\newcommand{\setCand}{\stySet{C}}

\newcommand{\defFunc}[3]{{{#1}\colon{#2}\to{#3}}}

\newcommand{\modifyColor}{black}
\newcommand{\modify}[1]{\textcolor{\modifyColor}{#1}}
\newcommand{\modifyTable}{\color{\modifyColor}}

\title{
One Single Hub Text Breaks CLIP: \\
Identifying Vulnerabilities in Cross-Modal Encoders via Hubness
}

\author{Hiroyuki Deguchi${}^\dagger$ ~~~ Katsuki Chousa${}^\dagger$ ~~~ Yusuke Sakai${}^\ddagger$\\
${}^\dagger$NTT, Inc. ~~~ ${}^\ddagger$Nara Institute of Science and Technology \\
\texttt{
\{hiroyuki.deguchi,katsuki.chousa\}@ntt.com
} \texttt{
sakai.yusuke.sr9@is.naist.jp
}}

\begin{document}
\maketitle
\begin{abstract}
The hubness problem, in which hub embeddings are close to many unrelated examples, occurs often in high-dimensional embedding spaces and may pose a practical threat for purposes such as information retrieval and automatic evaluation metrics.
In particular, since cross-modal similarity between text and images cannot be calculated by direct comparisons, such as string matching, cross-modal encoders that project different modalities into a shared space are helpful for various cross-modal applications, and thus, the existence of hubs may pose practical threats.
To reveal the vulnerabilities of cross-modal encoders, we propose a method for identifying the hub embedding and its corresponding hub text.
Experiments on image captioning evaluation in MSCOCO and nocaps along with image-to-text retrieval tasks in MSCOCO and Flickr30k showed that our method can identify a single hub text that unreasonably achieves comparable or higher similarity scores than human-written reference captions in many images, thereby revealing the vulnerabilities in cross-modal encoders.
\end{abstract}

\section{Introduction}
\label{sec:intro}

Cross-modal encoders such as CLIP~\citep{radford-etal-2021-learning,schuhmann-etal-2022-laionb,fang-etal-2024-data,chen-etal-2023-altclip}, which can calculate the semantic similarity between texts and images, are widely utilized for various applications, e.g., automatic evaluation metrics for caption quality and retrievers in cross-modal information retrieval, and have become fundamental technologies for multimodal processing.
Nevertheless, these models suffer from reliability issues due to the hubness problem~\citep{radovanovic-etal-2010-hubs}, in which a single example exhibits unreasonably high similarity scores with many irrelevant examples.
While several countermeasures for hubness in cross-modal encoders have been proposed~\citep{dinu-2015-etal-improving, lazaridou-etal-2015-hubness,huang-etal-2019-hubless,wang-etal-2023-balance, chowdhury-etal-2024-nearest}, concrete texts that are projected to hub embeddings have not yet been found.
In embedding-based automatic evaluation metrics for machine translation, \citet{deguchi-etal-2026-hacking} proposed a method for identifying a hub translation that is consistently evaluated as high quality across different inputs and references.
While this problem of single-modal encoders can be partially mitigated by combining string matching or other sanity checks, cross-modal similarity between text and images cannot be computed through direct comparisons and instead relies heavily on embeddings;
thus, the existence of hubs can deteriorate the reliability of cross-modal encoders.

\begin{figure}[t]
    \centering
    \includegraphics[width=\linewidth]{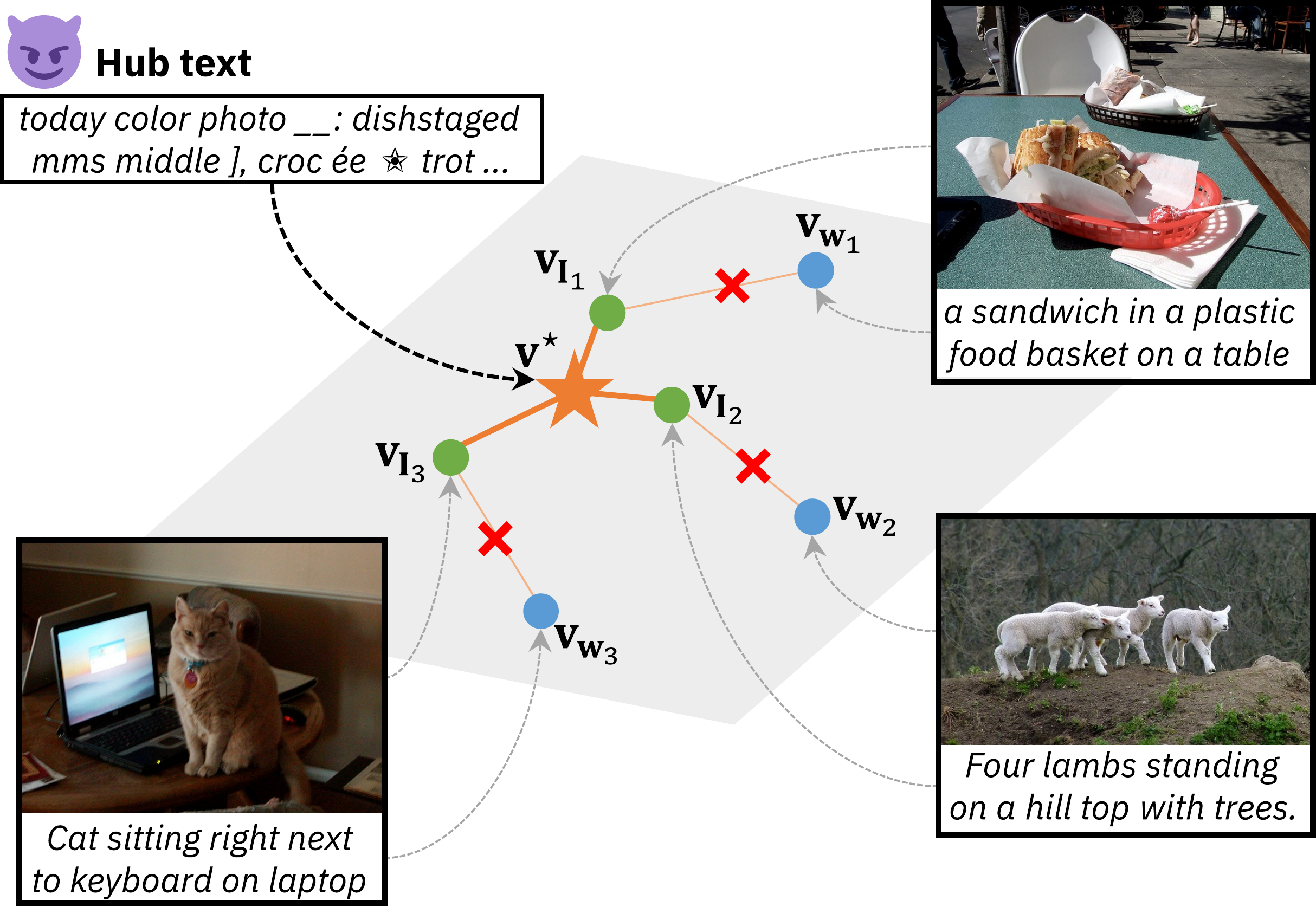}
    \caption{Hub text in cross-modal embedding space.
    Hub text has higher similarity with many unrelated images than human-written reference captions.
    }
    \label{fig:hubness}
\end{figure}

To reveal the vulnerabilities of cross-modal encoders,
we propose a method for identifying the hub text, which is a single text that exhibits unreasonably high similarity with many unrelated images.
\Cref{fig:hubness} illustrates the hubness problem and an example of hub text identified by our method in the embedding space of a cross-modal encoder.
Despite being semantically dissimilar, the hub text is embedded close to many unrelated images in the shared embedding space.
We first acquire an optimal hub embedding over a tuning dataset in the continuous embedding space.
Since cosine similarity, inner product, and squared Euclidean distance are commonly used to measure similarity in cross-modal tasks, we derive analytical solutions for an optimal hub embedding under these similarity measures.
We then decode the hub embedding using an inversion model that reconstructs input texts from their corresponding embeddings~\citep{morris-etal-2023-text}.
After decoding, we apply our beam local search to \modify{maximize similarity between} the hub text \modify{and multiple images}.
We iteratively replace each token in the decoded text with the token that maximizes the average similarity score across the tuning set, while considering multiple candidates.
Since our method operates as a black box, i.e., it works with only inputs and their corresponding embeddings, it can be applied to various models.

From our experiments, we confirmed that the proposed method can successfully identify the hub text for various cross-modal encoder models.
Specifically, the hub text identified with our method achieved a comparable or higher \metricClipscore{}~\citep{hessel-etal-2021-clipscore} than human reference captions and the hub text identified with the previous method in MSCOCO~\citep{lin-etal-2014-microsoft,karpathy-and-feifei-2015-deep} and nocaps~\citep{agrawal-etal-2019-nocaps}.
Moreover, we also observed that the contamination of hub text causes a significant drop in accuracy for image-to-text retrieval tasks in MSCOCO and Flickr30k~\citep{young-etal-2014-image}.
These findings reveal previously unexplored vulnerabilities in cross-modal encoders that may pose practical threats across a wide range of cross-modal applications, such as the evaluation metrics of \metricClipscore{} and cross-modal retrievers.

\section{Background and Related Work}
\label{sec:related}

\paragraph{Cross-modal encoder}
Cross-modal encoders, such as CLIP~\citep{radford-etal-2021-learning,schuhmann-etal-2022-laionb,fang-etal-2024-data,chen-etal-2023-altclip}, project texts and images into a shared embedding space, which enables calculating the similarity between them across modalities.
Their models are helpful and widely used for image-to-text retrieval and evaluation metrics of image captioning tasks, e.g., \metricClipscore{}~\citep{hessel-etal-2021-clipscore}.
We focus on models that have the same architecture as CLIP.

Let $\inputText{} \in \vocab^\ast$ and $\inputImage \in \spaceImage$ be a text and an image, respectively, where $\vocab^\ast$ is a Kleene closure of the vocabulary $\vocab$, and $\spaceImage \subseteq \R^{\sizeHeight \times \sizeWidth \times \sizeChannel}$ is a space of normalized images defined by a height $\sizeHeight \in \N$, width $\sizeWidth \in \N$, and number of channels $\sizeChannel \in \N$.
The cross-modal encoders $\defFunc{\funcEncoder_\paramEncoder}{ \spaceText \cup \spaceImage }{ \R^\sizeDim }$ project texts and images into their corresponding $\sizeDim{}~(\in \N)$-dimensional embeddings with learned parameters $\paramEncoder$.
By projecting them into the same space, we can calculate similarity across modalities.

\paragraph{CLIPScore}
\citet{hessel-etal-2021-clipscore} proposed \metricClipscore{}, an evaluation metric for image captioning tasks.
\metricClipscore{} evaluates the quality of a caption text $\inputText{}$ using the scaled cosine similarity between the caption text and its corresponding image $\inputImage$.
The evaluation score is calculated by the similarity function $\defFunc{\funcClipScore}{\R^\sizeDim \times \R^\sizeDim}{[0, \constScale]}$ with a scaling factor $\constScale \in \R$, as follows:
\begin{equation}
    \funcClipScore(\embedding{\inputText}, \embedding{\inputImage}) \coloneqq 
    \constScale \cdot \max\left(
    \frac{\embedding{\inputText}^\top \embedding{\inputImage}}{\lVert \embedding{\inputText} \rVert \lVert \embedding{\inputImage} \rVert }, 0\right),
    \label{eq:clipscore}
\end{equation}
where $\embedding{\bullet} \coloneqq \funcEncoder_\paramEncoder(\bullet)$.
In general, $\constScale = 2.5$ is used.
The corpus-level \metricClipscore{} is calculated by averaging scores for all caption--image pairs.

\paragraph{Hubness problem}
The hubness problem is a phenomenon in which hub embeddings in high-dimensional embedding spaces that are close to many other examples even though they are irrelevant~\citep{radovanovic-etal-2010-hubs}.
It causes unexpected behavior in tasks using embedding-based similarity, such as information retrieval and evaluation of text generation tasks, and reduces the reliability of the model.
Particularly in cross-modal tasks where direct comparisons of different modalities are impossible, e.g., string matching between two texts, reliance on embeddings is crucial, and several countermeasures for the hubness problem to mitigate its effects have been proposed~\citep{dinu-2015-etal-improving, lazaridou-etal-2015-hubness,huang-etal-2019-hubless,wang-etal-2023-balance, chowdhury-etal-2024-nearest}.
Even so, the underlying nature of hub embeddings has not been well studied.
Specifically, it remains unclear which specific examples are mapped to the hub embeddings, and whether such hubs are inherent to specific models or arise more generally across architectures and data distributions.

Some studies have identified such vulnerabilities in embedding models.
\citet{zhang-etal-2025-adversarial} generated adversarial images and audio that received high similarity scores with irrelevant examples in multimodal models.
Since images and audio are continuous representations, they can be easily obtained via gradient descent.

\paragraph{Data poisoning}
Reliable and safe data play a crucial role not only in model training data but also in retrieval-augmented generation~\citep{lewis-etal-2020-retrieval,karpukhin-etal-2020-dense,guu-etal-2020-retrieval,ma-etal-2023-query,ram-etal-2023-context,jeong-etal-2024-adaptive,an-etal-2025-rag}.
One vulnerability of recent large language models (LLMs) is their susceptibility to prompt-based attacks via data poisoning.
By injecting malicious or misleading examples into the retrieval data, attackers can manipulate the retrieved content and induce harmful or unintended model behaviors.
In response to these issues, recent studies have proposed various attack and defense methods, which suggests an increasing need for security in retrieval systems~\citep{hu-etal-2024-prompt,zou-etal-2025-poisonedrag,zhang-etal-2025-poisonedeye,tan-etal-2025-revprag,jiao-etal-2025-coordinated}.

\paragraph{Hub text identification}
Hub text identification is a more challenging task than identifying hub images and audio, as they are discrete representations.
In the most na\"ive method, we need to verify whether texts are mapped to hub embeddings for all possible texts, i.e., discrete token sequences, in $\vocab^\ast$ space, which is an NP-hard problem.
\citet{deguchi-etal-2026-hacking} proposed a 3-step approach for identifying the hub text, and revealed the existence of hub texts in \metricComet{}, a neural evaluation metric for translation quality.

They first trained a hub embedding that maximizes evaluation scores over the tuning data.
Since the scoring function in \metricComet{} is a non-linear feed-forward network, they used gradient descent for producing the hub embedding.
They then trained a hub decoder that generates concrete texts from their corresponding text embeddings, i.e., the inverse function of the text encoder, and decoded the hub embedding.
Finally, they refined the decoded text to maximize the score by using a greedy local search algorithm.
The algorithm sequentially finds the best token that maximizes the score over the tuning data for each token position, which is based on the greedy search.

\section{Proposed Method}
\label{sec:proposed}

To reveal vulnerabilities in cross-modal encoders and \metricClipscore{}, we propose a method for identifying hub text for their models.
We aim to identify a hub text that is close to arbitrary images:
\begin{align}
    \textHub^\star \coloneqq \argmax_{\textHub \in \spaceText} \sum_{\inputImage \in \spaceImage} \funcClipScore(
    \funcEncoder_\paramEncoder(\inputText)
    \funcEncoder_\paramEncoder(\inputImage)
    ).
\end{align}
Our method consists of three steps:
\textbf{(1) hub acquisition} analytically derives an optimal hub embedding in the embedding space, \textbf{(2) hub decoding} decodes the obtained hub embedding into its corresponding concrete text, and \textbf{(3) beam local search} refines the hub text to maximize the objective function with the proposed efficient algorithm.

\paragraph{(1) Hub acquisition}
We first acquire a hub embedding $\embedding{}^\star \in \R^\sizeDim$, which is close to arbitrary image embeddings, by maximizing the objective
\begin{align}
    \funcObj(\embedding{}; \setDataImage) \coloneqq \frac{1}{|\setDataImage|} \sum_{\inputImage \in \setDataImage} \funcClipScore(
    \embedding{},
    \embedding{\inputImage}
    ),
\end{align}
where $\setDataImage \subset \spaceImage$ is a tuning data that consists of multiple images.
In contrast to prior work, which relied on gradient descent to obtain the hub embedding because the COMET scoring function is a non-linear feed-forward network, we derive an analytical solution utilizing the nature of the scoring function $\funcClipScore$ in \metricClipscore{}.
The optimal solution can be obtained simply by averaging all normalized image embeddings over the tuning data, as follows:
\begin{equation}
    \!\embedding{}^\star \coloneqq \argmax_{\embedding{} \in \R^\sizeDim} \funcObj(\embedding{}; \setDataImage) 
    = \frac{1}{|\setDataImage|} \sum_{\inputImage \in \setDataImage} \frac{\embedding{\inputImage}}{\lVert \embedding{\inputImage} \rVert}.\!
\end{equation}
This solution is derived from the equality condition of the Cauchy--Schwarz inequality.
We further provide the detailed derivation of an optimal hub embedding on not only cosine similarity but also other widely used similarity functions, inner product, and squared Euclidean distance, in \cref{sec:derive}.

\begin{algorithm}[t]
  \small
  \caption{Beam local search}
  \label{alg:lbs}
  \Given{Scoring function
  $\defFunc{\funcClipScore}{\R^\sizeDim \times \R^\sizeDim}{[0, \constScale]}$ 
  and function $\funcTopk\colon 2^{\spaceText \times [0, \constScale]} \to
  \{\setBeam \mid \setBeam \subseteq 2^{\spaceText \times [0, \constScale]} \land |\setBeam| = k \}
  $ that returns the top-$k~(\in \N)$ candidates.
  }
  \Input{Initial hub text candidate $\textHub^{\text{init}} \in \spaceText$ and tuning data $\setDataImage \subset \spaceImage$.} 
  \Output{Hub text $\textHub^\star \in \spaceText$.}

  $t \gets 0$, 
  $\setBeam^{(0)} \gets \{ (\textHub^{\text{init}}, 
  \funcObj( \funcEncoder_\paramEncoder(\textHub^{\text{init}}); \setDataImage ) )
  \}$ \;
  Initialize a new HashMap $\bm{\mathcal{P}}\colon \spaceText \to 2^\N$ \label{alg:lbs:hashmap} \;
  $\bm{\mathcal{P}}(\textHub^{\text{init}}) \gets \{ 1, \ldots, |\textHub^{\text{init}}| \}$\;
  \Repeat{
    $\forall (\textHub, \score) \in \setBeam^{(t)}. \bm{\mathcal{P}}(\textHub) = \varnothing$
    \label{alg:lbs:convergence}
  }{
    $t \gets t + 1$\;
    $\setCand \gets \setBeam^{(t-1)}$\;
    \Foreach{$(\textHub, \score) \in \setBeam^{(t-1)}$}{
        \If{$\bm{\mathcal{P}}(\textHub) = \varnothing$}{\Continue}
        $i \sim \bm{\mathcal{P}}(\textHub) $\label{alg:lbs:order} \;
        {
          \Foreach{$v \in \vocab$ \label{alg:lbs:vocabloop}}{
            \tcp{$\circ$ denotes concatenation.}
            \tcp{$\textHub_{a:b}$ denotes the slice of  $\textHub$ from $a$ to $b$ inclusive.
            }
            $\textHub^\text{cand} \gets \textHub_{1:i-1} \circ v \circ \textHub_{i+1:|\textHub|}$\;
            $\score^\text{cand} \gets \funcObj(\funcEncoder_\paramEncoder(\textHub^\text{cand}); \setDataImage)$\;
            $\setCand \gets \setCand \cup \{ (\textHub^\text{cand}, \score^\text{cand}) \}$\;
          }
        }
        $\bm{\mathcal{P}}(\textHub) \gets \bm{\mathcal{P}}(\textHub) \setminus \{ i \}$\;
    }
    $\setBeam^{(t)} \gets \funcTopk(\setCand)$ \label{alg:lbs:topk} \;
    \If{$\setBeam^{(t)} \neq \setBeam^{(t-1)}$}{
        Initialize a new HashMap $\bm{\mathcal{P}}\colon \spaceText \to 2^\N$\;
        \Foreach{$(\textHub, \score) \in \setBeam^{(t)}$}{
            $\bm{\mathcal{P}}(\textHub) \gets \{ 1, \ldots, |\textHub| \}$\;
        }
    }
  }
  \Return{$\displaystyle\argmax_{\textHub:(\textHub, \score) \in \setBeam^{(t)}} \score$}
\end{algorithm}

\paragraph{(2) Hub decoding}
Next, we decode the hub embedding into its corresponding concrete text.
Since most cross-modal encoders $f$ are non-linear functions and their inverse functions $f^{-1}$ cannot be calculated exactly, we train an inversion model $\paramInverter$ that reconstructs input texts from their embeddings~\citep{morris-etal-2023-text}.
It is trained by minimizing the following negative log-likelihood loss:
\begin{align}
    \funcLoss(\paramInverter; \setDataText) &\coloneqq -\sum_{\inputText \in \setDataText}
    \log p_\phi(\inputText | \funcEncoder_\paramEncoder(\inputText)), \\
    \hat{\paramInverter} &\coloneqq \argmin_{\paramInverter} \funcLoss(\paramInverter; \setDataText),
\end{align}
where $\setDataText \subset \spaceText$ is a training data for the inversion model.
During decoding, we generate multiple hypotheses of the hub text $\setHypotheses \subset \spaceText$ from the hub embedding $\embedding{}^\star$, obtained in Step (1), using the trained inversion model $\hat{\paramInverter}$:
\begin{align}
\setHypotheses \coloneqq \left\{ \inputText_i \right\}_{i=1}^{|\setHypotheses|}, ~~~
\inputText_i \sim p_{\hat{\phi}}(\inputText | \embedding{}^\star).
\end{align}
Then, we select the best hypothesis that maximizes the objective function $\funcObj$ over the tuning data $\setDataImage$:
\begin{align}
    \argmax_{\inputText \in \setHypotheses} \funcObj(\funcEncoder_\paramEncoder(\inputText); \setDataImage).
\end{align}

\paragraph{(3) Beam local search}
Finally, we refine the decoded hub text to maximize the score.
Our search algorithm (\cref{alg:lbs}) receives the initial hub text, which is the decoded hub text obtained by Step (2), and the main loop iteratively replaces a token in each candidate sequence with a token that maximizes the corpus-level score.
In \cref{alg:lbs:hashmap}, we introduce a hash map $\bm{\mathcal{P}}$ to manage search states and check for convergence.
The algorithm terminates when no token updates occur for any candidate in the beam, as specified by the stopping criterion in \cref{alg:lbs:convergence}.
This design enables a more flexible search.
While prior work sequentially replaces each token in the decoded text in a left-to-right manner, our method allows tokens to be replaced in a random order (\cref{alg:lbs:order}).
The most significant improvement over the conventional method is the extension from greedy search to beam search, which maintains multiple candidates and enables a more efficient search of a larger space.

\section{Experiments}
\label{sec:exp}

\begin{table*}[t]
    \centering
    \small
    \tabcolsep 5pt
    \begin{tabular}{@{}l rrrr rrrr@{}}
        \toprule
        & \multicolumn{4}{c}{MSCOCO \modify{(in-domain)}} & \multicolumn{4}{c}{nocaps \modify{(out-of-domain)}} \\ \cmidrule(lr){2-5} \cmidrule(l){6-9}
        & \multicolumn{2}{c}{Caption text} & \multicolumn{2}{c}{Hub text} 
        & \multicolumn{2}{c}{Caption text} & \multicolumn{2}{c}{Hub text} \\
        \cmidrule(lr){2-3} \cmidrule(lr){4-5}
        \cmidrule(lr){6-7} \cmidrule(l){8-9}
        Model
        & BLIP-2 & Human & GLS & Ours
        & BLIP-2 & Human & GLS & Ours
        \\
        \toprule
\texttt{openai/clip-vit-base-patch32} & 0.739 & \underline{0.759} & 0.732 & ${}^{\dagger}$\textbf{0.842} & 0.740 & \underline{0.758} & 0.700 & ${}^{\dagger}$\textbf{0.814} \\
\texttt{openai/clip-vit-large-patch14} & 0.613 & \underline{0.639} & 0.633 & ${}^{\dagger}$\textbf{0.649} & 0.608 & \textbf{0.623} & 0.612 & ${}^{\dagger}$\underline{0.622} \\
\texttt{openai/clip-vit-large-patch14-336} & 0.628 & 0.654 & \underline{0.677} & ${}^{\dagger}$\textbf{0.701} & 0.616 & \underline{0.631} & 0.620 & ${}^{\dagger}$\textbf{0.663} \\
\texttt{laion/CLIP-ViT-L-14-laion2B-s32B-b82K} & 0.720 & \underline{0.748} & 0.690 & ${}^{\dagger}$\textbf{0.782} & 0.719 & \textbf{0.740} & 0.592 & ${}^{\dagger}$\underline{0.729} \\
\texttt{laion/CLIP-ViT-H-14-laion2B-s32B-b79K} & 0.729 & 0.771 & \underline{0.78}0 & ${}^{\dagger}$\textbf{0.825} & \underline{0.728} & \textbf{0.759} & 0.679 & ${}^{\dagger}$0.716 \\
\texttt{laion/CLIP-ViT-g-14-laion2B-s12B-b42K} & 0.688 & 0.721 & \textbf{0.767} & \underline{0.747} & \underline{0.689} & \textbf{0.714} & 0.666 & ${}^{\dagger}$0.680 \\
\texttt{apple/DFN2B-CLIP-ViT-L-14} & 0.678 & 0.722 & \underline{0.723} & ${}^{\dagger}$\textbf{0.814} & 0.685 & \underline{0.712} & 0.646 & ${}^{\dagger}$\textbf{0.737} \\
\texttt{apple/DFN5B-CLIP-ViT-H-14} & 0.789 & 0.838 & \underline{0.965} & ${}^{\dagger}$\textbf{0.974} & 0.809 & \underline{0.837} & 0.831 & ${}^{\dagger}$\textbf{0.841} \\
\texttt{apple/DFN5B-CLIP-ViT-H-14-378} & 0.777 & 0.837 & \underline{0.995} & ${}^{\dagger}$\textbf{1.023} & \underline{0.814} & \textbf{0.841} & 0.798 & ${}^{\dagger}$\underline{0.814} \\
\texttt{BAAI/AltCLIP} & 0.622 & \underline{0.635} & 0.512 & ${}^{\dagger}$\textbf{0.643} & 0.622 & \underline{0.623} & 0.479 & ${}^{\dagger}$\textbf{0.628} \\
\bottomrule
    \end{tabular}
    \caption{
    \metricClipscore{} evaluated by various models. 
    ``Human'' indicates reference captions.
    To evaluate corpus-level scores, each hub text is repeated to match the size of test set, as it is single text.
    Best and second-best scores for each dataset are indicated in \textbf{bold font} and \underline{underline}, respectively.
    \modify{
    ``$\dagger$'' denotes scores where our identified hub text statistically outperformed ``GLS'' ($p<0.05$).
    }
    }
    \label{tab:clips:corpus_scores}
\end{table*}

\begin{table*}[t]
    \centering
    \tabcolsep 3.5pt
    \begin{tabular}{@{}lr p{9.6cm} c@{}}
    \toprule
    Caption & \metricClipscore{} & Text & Image \\
    \midrule
    BLIP-2 & 0.750 & two boys are skateboarding down a street with their skateboards
    & \multirow{4}{*}{\includegraphics[height=1.8cm]{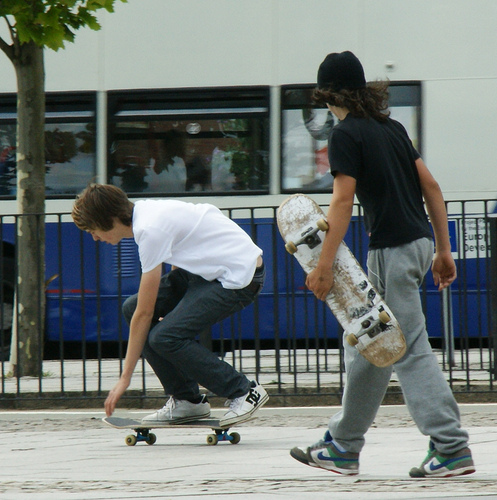}}
    \\
    Human & 0.793 & A couple of young boys with skateboards pass a city bus \\
    Hub & 1.012 & today color photo \_\_: dishstaged mms middle ], croc ée \ding{76} trot maker gely bw 8 boarded\verb|<U+FE0F>|: garethapproached cision \\
    \midrule
    BLIP-2 & 0.652 & a cat sitting on a table
    & \multirow{4}{*}{  \includegraphics[height=1.8cm,width=1.8cm]{fig/COCO_val2014_000000055466.jpg}}
    \\
    Human & 0.780 & Cat sitting right next to keyboard on laptop \\
    Hub & 0.981 & today color photo \_\_: dishstaged mms middle ], croc ée \ding{76} trot maker gely bw 8 boarded\verb|<U+FE0F>|: garethapproached cision \\
    \midrule
    BLIP-2 & 0.797 & a small kitchen with a stove and microwave oven & \multirow{4}{*}{  \includegraphics[height=1.8cm,width=1.8cm]{fig/COCO_val2014_000000036082.jpg}} \\
    Human & 0.806 & a kitchen with a small refrigerator and a microwave oven \\
    Hub & 1.034 & today color photo \_\_: dishstaged mms middle ], croc ée \ding{76} trot maker gely bw 8 boarded\verb|<U+FE0F>|: garethapproached cision \\
    \bottomrule
    \end{tabular}
    \caption{
    Example scores of captions and the single hub text in \texttt{openai/clip-vit-base-patch32} on MSCOCO
    }
    \label{tab:case:mscoco:test}
\end{table*}

\begin{table*}[t]
    \centering
    \small
    \tabcolsep 3pt
    \begin{tabularx}{\linewidth}{@{}lX@{}}
    \toprule
    Model & Hub text \\
    \midrule
    \texttt{openai/clip-vit-base-patch32} & today color photo \_\_: dishstaged mms middle ], croc ée \ding{76} trot maker gely bw 8 boarded\verb|<U+FE0F>|: garethapproached cision \\
\texttt{openai/clip-vit-large-patch14} & photo taken using dnskarchivesdgs unparalleled \\
\texttt{openai/clip-vit-large-patch14-336} & degrees photographer , " toc more " av benefchu (- tely his latest ' buenas wiscondged kirby pa (@ \\
\bottomrule
    \end{tabularx}
    \caption{Examples of hub texts}
    \label{tab:hubtexts}
\end{table*}

\subsection{Hub text identification}
We identified and evaluated hub texts of CLIP~\citep{radford-etal-2021-learning}, LION-CLIP~\citep{schuhmann-etal-2022-laionb}, DFN-CLIP~\citep{fang-etal-2024-data}, and AltCLIP~\citep{chen-etal-2023-altclip} in image captioning and image--text retrieval tasks.
For all models, we used the validation set of the MSCOCO dataset~\citep{lin-etal-2014-microsoft,karpathy-and-feifei-2015-deep} for the tuning data $\setDataImage$.
In all analyses, if no model name is indicated, we used \texttt{openai/clip-vit-base-patch32}.
We compared the single hub text identified by our method (Ours) with that identified by the sequential greedy local search (GLS) proposed by \citet{deguchi-etal-2026-hacking}.

\paragraph{Inversion model training}
We trained inversion models for each cross-modal encoder from mT5-base\footnote{\texttt{google/mt5-base}}~\citep{xue-etal-2021-mt5}, a pretrained encoder-decoder model, with frozen text embeddings encoded by each cross-modal encoder.
For the training data $\setDataText$, we used the caption texts in the MSCOCO training set.
The inversion models were optimized using the AdamW optimizer ($\beta_1=0.9$, $\beta_2=0.999$, $\epsilon=10^{-8}$)~\citep{loshchilov-etal-2018-decoupled} with a learning rate of $3 \times 10^{-4}$, 4,000 warm-up steps, and training for 20 epochs using a batch size of 128 captions.

\paragraph{Hub decoding}
During decoding, we generated 4,096 caption hypotheses from a single hub embedding obtained by Step (1) and then selected the best one using the tuning data.
To diversify the hypotheses, we used epsilon sampling with $\varepsilon=0.02$~\citep{hewitt-etal-2022-truncation,freitag-etal-2023-epsilon}.
\modify{
Note that this step does not require expensive computational resources because the decoding cost is equivalent to generating 4,096 sentences with mT5-base, which takes less than 1 minute on a single GPU.
}

\paragraph{Beam local search}
We applied beam local search to the decoded texts with multiple beam sizes $k \in \{5, 10, 20\}$ and selected the best result on the tuning data.
For \texttt{openai/clip-vit-base-patch32}, the beam local search replaced tokens 382 times with a sequence length of 23 tokens.
It took 12,486 seconds on 8 NVIDIA RTX\trademark{} 6000Ada GPUs.

\subsection{Evaluation in image captioning}

\begin{table*}[t]
    \centering
    \small
    \tabcolsep 2.8pt
    \begin{NiceTabular}{@{}r rrrrrrrrrr rrrrrrrrrr@{}}
    \CodeBefore
    \rowcolor{gray!20}{4,8,12,16,20}
    \Body
    \toprule
    & \multicolumn{10}{@{}c@{}}{MSCOCO} & \multicolumn{10}{@{}c@{}}{Flickr30k} \\
    \cmidrule(lr){2-11} \cmidrule(l){12-21}
    & \multicolumn{2}{@{}c@{}}{NDCG} & \multicolumn{2}{@{}c@{}}{MAP} & \multicolumn{2}{@{}c@{}}{Recall} & \multicolumn{2}{@{}c@{}}{Precision} & \multicolumn{2}{@{}c@{}}{MRR} 
    & \multicolumn{2}{@{}c@{}}{NDCG} & \multicolumn{2}{@{}c@{}}{MAP} & \multicolumn{2}{@{}c@{}}{Recall} & \multicolumn{2}{@{}c@{}}{Precision} & \multicolumn{2}{@{}c@{}}{MRR} 
    \\
    \cmidrule(lr){2-3} \cmidrule(lr){4-5} \cmidrule(lr){6-7} \cmidrule(lr){8-9} \cmidrule(lr){10-11}
    \cmidrule(lr){12-13} \cmidrule(lr){14-15} \cmidrule(lr){16-17} \cmidrule(lr){18-19} \cmidrule(l){20-21}
    \#CT 
    & ${}^\text{@}$1 & ${}^\text{@}$10 & ${}^\text{@}$1 & ${}^\text{@}$10 & ${}^\text{@}$1 & ${}^\text{@}$1k & ${}^\text{@}$1 & ${}^\text{@}$5 & ${}^\text{@}$1 & ${}^\text{@}$10 
    & ${}^\text{@}$1 & ${}^\text{@}$10 & ${}^\text{@}$1 & ${}^\text{@}$10 & ${}^\text{@}$1 & ${}^\text{@}$1k & ${}^\text{@}$1 & ${}^\text{@}$5 & ${}^\text{@}$1 & ${}^\text{@}$10 
    \\
    \midrule
\multicolumn{21}{@{}l@{}}{\textbf{\texttt{openai/clip-vit-base-patch32}}} \\
0 & 50.5 & 44.3 & 10.1 & 33.0 & 10.1 & 99.0 & 50.5 & 34.2 & 50.5 & 60.9 & 78.0 & 71.6 & 15.6 & 60.2 & 15.6 & 99.8 & 78.0 & 58.2 & 78.1 & 85.4 \\
1 & 44.2 & 42.8 & 8.8 & 31.4 & 8.8 & 99.0 & 44.2 & 33.3 & 44.2 & 57.2 & 70.0 & 68.8 & 14.0 & 56.7 & 14.0 & 99.8 & 70.0 & 55.8 & 70.0 & 80.5 \\
1,000 & 44.1 & 35.1 & 8.8 & 26.3 & 8.8 & 52.2 & 44.1 & 27.5 & 44.1 & 51.5 & 70.1 & 56.0 & 14.0 & 46.6 & 14.0 & 58.6 & 70.1 & 46.1 & 70.1 & 74.4 \\
\midrule
\multicolumn{21}{@{}l@{}}{\textbf{\texttt{openai/clip-vit-large-patch14-336}}} \\
0 & 57.0 & 49.7 & 11.4 & 38.2 & 11.4 & 99.3 & 57.0 & 38.9 & 57.0 & 66.8 & 87.6 & 79.8 & 17.5 & 69.9 & 17.5 & 99.9 & 87.6 & 67.0 & 87.6 & 92.0 \\
1 & 52.4 & 49.0 & 10.5 & 37.1 & 10.5 & 99.3 & 52.4 & 38.3 & 52.4 & 64.2 & 76.5 & 76.1 & 15.3 & 65.0 & 15.3 & 99.9 & 76.5 & 63.3 & 76.5 & 86.1 \\
1,000 & 52.4 & 40.7 & 10.5 & 31.1 & 10.5 & 52.9 & 52.4 & 32.0 & 52.4 & 59.3 & 76.5 & 60.7 & 15.3 & 51.8 & 15.3 & 60.7 & 76.5 & 51.1 & 76.5 & 79.5 \\
\midrule
\multicolumn{21}{@{}l@{}}{\textbf{\texttt{laion/CLIP-ViT-g-14-laion2B-s12B-b42K}}} \\
0 & 64.7 & 57.9 & 13.0 & 46.5 & 13.0 & 99.6 & 64.7 & 46.3 & 64.7 & 73.5 & 91.4 & 85.4 & 18.3 & 77.5 & 18.3 & 99.9 & 91.4 & 73.6 & 91.4 & 94.8 \\
1 & 59.7 & 55.9 & 11.9 & 44.2 & 11.9 & 99.7 & 59.7 & 44.6 & 59.7 & 70.4 & 88.6 & 83.9 & 17.7 & 75.3 & 17.7 & 99.9 & 88.6 & 71.4 & 88.6 & 93.3 \\
1,000 & 59.8 & 47.5 & 12.0 & 37.9 & 12.0 & 54.5 & 59.8 & 38.6 & 59.8 & 66.2 & 88.6 & 76.4 & 17.7 & 68.3 & 17.7 & 76.3 & 88.6 & 66.9 & 88.6 & 91.6 \\
\midrule
\multicolumn{21}{@{}l@{}}{\textbf{\texttt{apple/DFN5B-CLIP-ViT-H-14-378}}} \\
0 & 70.4 & 63.1 & 14.1 & 51.9 & 14.1 & 99.7 & 70.4 & 51.0 & 70.4 & 78.4 & 92.2 & 87.9 & 18.4 & 81.0 & 18.4 & 100.0 & 92.2 & 77.2 & 92.2 & 95.2 \\
1 & 41.1 & 55.6 & 8.2 & 43.0 & 8.2 & 99.8 & 41.1 & 46.4 & 41.1 & 62.0 & 63.4 & 79.1 & 12.7 & 68.6 & 12.7 & 100.0 & 63.4 & 68.5 & 63.4 & 80.1 \\
1,000 & 41.0 & 26.8 & 8.2 & 21.1 & 8.2 & 24.2 & 41.0 & 21.5 & 41.0 & 43.0 & 63.5 & 41.8 & 12.7 & 34.6 & 12.7 & 35.6 & 63.5 & 34.6 & 63.5 & 64.3 \\
\midrule
\multicolumn{21}{@{}l@{}}{\textbf{\texttt{BAAI/AltCLIP}}} \\
0 & 57.9 & 52.0 & 11.6 & 40.6 & 11.6 & 99.5 & 57.9 & 41.1 & 57.9 & 67.8 & 85.5 & 80.9 & 17.1 & 71.8 & 17.1 & 99.9 & 85.5 & 68.3 & 85.5 & 90.8 \\
1 & 50.4 & 49.5 & 10.1 & 37.6 & 10.1 & 99.5 & 50.4 & 38.8 & 50.4 & 63.3 & 69.6 & 75.8 & 13.9 & 64.6 & 13.9 & 99.9 & 69.6 & 62.4 & 69.6 & 82.2 \\
1,000 & 50.5 & 38.4 & 10.1 & 29.3 & 10.1 & 49.3 & 50.5 & 30.2 & 50.5 & 57.0 & 69.9 & 52.4 & 14.0 & 43.8 & 14.0 & 49.6 & 69.9 & 43.7 & 69.9 & 72.9 \\

    \bottomrule
    \end{NiceTabular}
    \caption{
    Results of image-to-text retrieval tasks with hub text contamination in MSCOCO and Flickr30k.
    \#CT is the number of contaminations, which denotes the number of insertions of the single hub text into the document index.
    }
    \label{tab:i2t:results}
\end{table*}

\paragraph{Setup}
Our aim here is to investigate whether a hub text is evaluated with an unreasonably high score in image captioning tasks.
To validate the effectiveness of our method, we compared the single hub text and caption texts for each image.
Specifically, we evaluated for caption texts generated by BLIP-2-FlanT5-XL\footnote{
\texttt{Salesforce/blip2-flan-t5-xl}
}~\citep{li-etal-2023-blip-2,chung-etal-2024-scaling}, reference captions created by humans, and a single hub text generated by GLS~\citep{deguchi-etal-2026-hacking} and by the proposed method.
For corpus-level evaluation, we repeated the hub text to match the size of the test set, as it is a single text.
We used the test set of the MSCOCO and the validation set\footnote{
The reference captions are available only in the validation set.
}
of the nocaps dataset~\citep{agrawal-etal-2019-nocaps} for evaluation.
\modify{
We compared our method with GLS using statistical significance tests via paired bootstrap resampling with 1,000 resamples~\citep{koehn-2004-statistical}.
}

\paragraph{Results}

\Cref{tab:clips:corpus_scores} lists the \metricClipscore{} calculated by various cross-modal encoder models.
Despite evaluating a single hub text, which is semantically irrelevant to the input images, our method achieved higher scores than reference captions created by humans, image-by-image, on many models.
In addition, our method achieved higher scores than the previous method, GLS, in most models.
Especially in nocaps, i.e., the out-of-domain dataset, the \metricClipscore{} of GLS is lower than that of reference captions, whereas our method outperformed it in five models.
Furthermore, we confirmed that our method statistically outperformed GLS in 9 of 10 models on MSCOCO and all models on nocaps, demonstrating that our method identifies hub texts more effectively than GLS.

We present the concrete hub text identified by our method and its \metricClipscore{} in \cref{tab:case:mscoco:test}.
The hub text is a single string unrelated to any images, yet it achieves unreasonably higher scores than human reference captions.
Interestingly, the hub text contains terms such as ``color'' and ``photo'', which are presumably likely to appear frequently in the training data of CLIP.
Other examples of hub texts are listed in \cref{tab:hubtexts}.
The hub texts for other models also contain ``photo'' or ``photographer''.
These findings suggest that the hub text may occur due to the training data distribution.

\subsection{Image-to-text retrieval}

\begin{table}[t]
    \centering
    \small
    \modifyTable
    \tabcolsep 2pt
    \begin{NiceTabular}{@{}l rr rr rr rr rr@{}}
    \CodeBefore
    \rowcolor{gray!20}{4,7}
    \Body
    \toprule
    & \multicolumn{2}{@{}c@{}}{NDCG} & \multicolumn{2}{@{}c@{}}{MAP} & \multicolumn{2}{@{}c@{}}{Recall} & \multicolumn{2}{@{}c@{}}{Precision} & \multicolumn{2}{@{}c@{}}{MRR} \\
    \cmidrule(lr){2-3} \cmidrule(lr){4-5} \cmidrule(lr){6-7} \cmidrule(lr){8-9} \cmidrule(l){10-11}
    \#CT & ${}^\text{@}$1 & ${}^\text{@}$10 & ${}^\text{@}$1 & ${}^\text{@}$10	& ${}^\text{@}$1 & ${}^\text{@}$1k & ${}^\text{@}$1 & ${}^\text{@}$5 & ${}^\text{@}$1 & ${}^\text{@}$10 \\
    \midrule
    0  & 50.5 & 44.3 & 10.1 & 33.0 & 10.1 & 99.0 & 50.5 & 34.2 & 50.5 & 60.9 \\
    \midrule
    \multicolumn{11}{@{}l@{}}{
    Injected text: Randomly selected caption text
    } \\
    1 & 50.5 & 44.4 & 10.1 & 33.0 & 10.1 & 99.0 & 50.5 & 34.2 & 50.5 & 60.9 \\
    1,000 & 50.5 & 44.4 & 10.1 & 33.0 & 10.1 & 98.6 & 50.5 & 34.2 & 50.5 & 60.9 \\
    \midrule
    \multicolumn{11}{@{}l@{}}{
    Injected text: Single hub text
    } \\
    1 & 44.2 & 42.8 & 8.8 & 31.4 & 8.8 & 99.0 & 44.2 & 33.3 & 44.2 & 57.2 \\
    1,000 & 44.1 & 35.1 & 8.8 & 26.3 & 8.8 & 52.2 & 44.1 & 27.5 & 44.1 & 51.5 \\
    \bottomrule
    \end{NiceTabular}
    \caption{\modify{Comparisons of image-to-text retrieval performance on MSCOCO with \texttt{openai/clip-vit-base-} \texttt{patch32} under random caption and hub text injection}}
    \label{tab:i2t:results:random}
\end{table}

\paragraph{Setup}
We experimented with image-to-text (I2T) retrieval tasks on MSCOCO and Flickr30k~\citep{young-etal-2014-image} using MTEB~\citep{muennighoff-etal-2023-mteb}.
Assuming attackers contaminate the documents to be searched, we insert the single hub text for each model into the document side.
We also simulated attack cases where the single hub text is repeated and inserted multiple times for search engine optimization (SEO) or cracking.
The retrieval performance was evaluated using the normalized discounted cumulative gain (NDCG) at top-1 and top-10, mean average precision (MAP) at top-1 and top-10, recall at top-1 and top-1,000, precision at top-1 and top-5, and mean reciprocal rank (MRR) at top-1 and top-10.

\paragraph{Results}
The results of the I2T retrieval tasks are demonstrated in \cref{tab:i2t:results}.
Here, ``\#CT'' denotes the number of contaminations, which means the number of insertions of the single hub text into the document index.
``\#CT = 0'' is the baseline retrieval performance of each model.
We observed that all metrics at top-1 significantly degraded, regardless of the model, even though just a single text was inserted, i.e., when \#CT = 1.
In particular, precision at top-1 decreased by up to 29.3\%, which means that the top-1 search result is often contaminated by the hub text, even though relevant texts are included in the search index.
\modify{
We also evaluated the retrieval performance at \#CT = 1,000, which corresponds to a different scenario rather than a single-injection setting.
Our intention here is to approximate large-scale duplication of similar content for exploiting search engines and recommendation systems, e.g., template-based generation and spam replication.
}
In this setting, recall at top-1,000 significantly decreased by up to 75.5\%.
This indicates that contamination using the hub text could hinder the retrieval of relevant texts.

\modify{
To verify that the observed performance degradation was due to the hub text, we also compared the retrieval performance when a caption randomly selected from the training data was injected.
\cref{tab:i2t:results:random} shows performance comparisons on MSCOCO for random caption injection and hub text injection.
Injecting a random caption did not degrade performance, whereas injecting the hub text did.
These results indicate that hub texts, unlike other captions, pose a potential threat.
}

\section{Discussion}
\label{sec:discussion}

\subsection{Statistics of instance-level scores}

\begin{table}[t]
    \centering
    \footnotesize
    \tabcolsep 2.5pt
    \begin{tabular}{@{}l rrrr @{}}
    \toprule
          & \multicolumn{4}{@{}c@{}}{Win rate: Hub > Human} \\
          \cmidrule(l){2-5}
          & \multicolumn{2}{@{}c@{}}{MSCOCO} & \multicolumn{2}{@{}c@{}}{nocaps} \\
          \cmidrule(lr){2-3} \cmidrule(l){4-5}
    Model & GLS & Ours & GLS & Ours \\
    \midrule
\scriptsize\texttt{openai/clip-vit-base-patch32} & 39.1 & 78.6 & 27.5 & 71.1 \\
\scriptsize\texttt{openai/clip-vit-large-patch14} & 48.4 & 54.7 & 45.2 & 49.5 \\
\scriptsize\texttt{openai/clip-vit-large-patch14-336} & 60.1 & 67.7 & 47.4 & 62.6 \\
\scriptsize\texttt{laion/CLIP-ViT-L-14-laion2B-s32B-b82K} & 33.4 & 60.1 & 14.2 & 44.3 \\
\scriptsize\texttt{laion/CLIP-ViT-H-14-laion2B-s32B-b79K} & 52.8 & 65.9 & 26.2 & 35.4 \\
\scriptsize\texttt{laion/CLIP-ViT-g-14-laion2B-s12B-b42K} & 63.4 & 58.2 & 35.2 & 38.4 \\
\scriptsize\texttt{apple/DFN2B-CLIP-ViT-L-14} & 51.4 & 76.6 & 29.6 & 56.8 \\
\scriptsize\texttt{apple/DFN5B-CLIP-ViT-H-14} & 81.8 & 83.9 & 48.0 & 51.1 \\
\scriptsize\texttt{apple/DFN5B-CLIP-ViT-H-14-378} & 87.3 & 90.0 & 37.5 & 41.1 \\
\scriptsize\texttt{BAAI/AltCLIP} & 12.9 & 52.1 & 9.0 & 51.8 \\
    \bottomrule
    \end{tabular}
    \caption{
    Instance-level win rates (\%) in \metricClipscore{} compared with human reference captions
    }
    \label{tab:clips:winrate}
\end{table}

\begin{figure}[t]
    \centering
    \includegraphics[width=\linewidth]{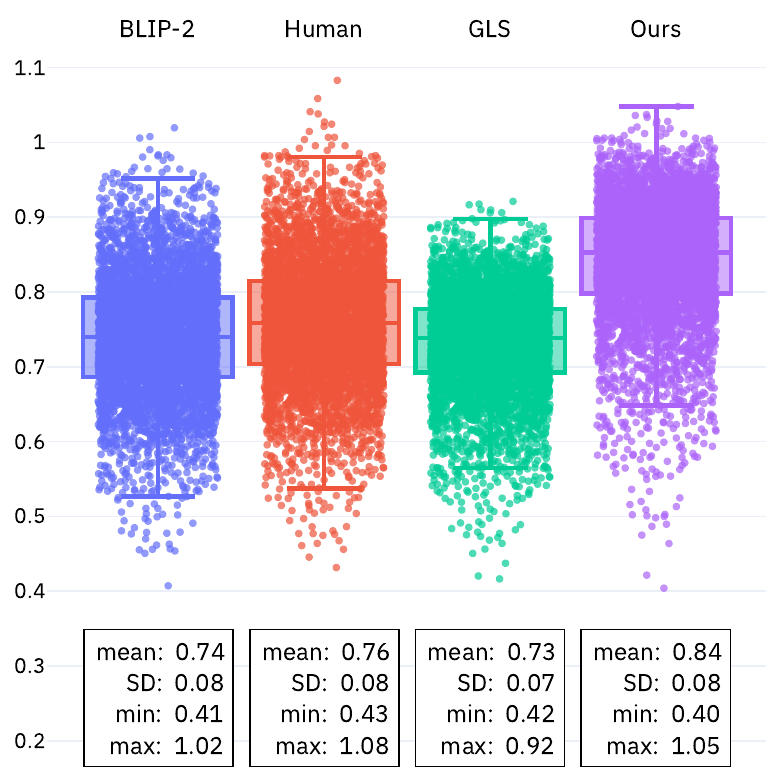}
    \caption{
    Scatter plots of instance-level scores.
    Our identified single hub text achieved higher scores than BLIP-2, Human, and GLS in many instances.
    }
    \label{fig:clips:scatter}
\end{figure}

We investigated the statistics of instance-level scores in the captioning evaluation experiments.
\Cref{tab:clips:winrate} shows the instance-level win rates in \metricClipscore{} compared with human reference captions.
The single hub text found with our method achieved higher scores than human reference captions in many models.
For both datasets, our proposed method identified more influential hub texts than conventional methods in most models.
Specifically, in MSCOCO, our found hub text outperformed reference captions in more than half of the test cases.
Furthermore, even models that have high correlations with human assessments~\citep{gomes-etal-2025-evaluation}, \texttt{apple/DFN5B-CLIP-ViT-H-14-378}, received higher scores than the reference captions in 90\% of test cases, suggesting that hub texts exist regardless of the model performance.

We also compared the instance-level scores between captions and hub texts.
\Cref{fig:clips:scatter} shows scatter plots of the instance-level scores in MSCOCO.
Human reference captions achieved higher scores than captions generated using BLIP-2 and the single hub text identified with GLS, while the hub text identified with our method significantly outperformed the reference captions.

\subsection{Trajectory in local search}

\begin{figure}[t]
    \centering
    \includegraphics[width=\linewidth]{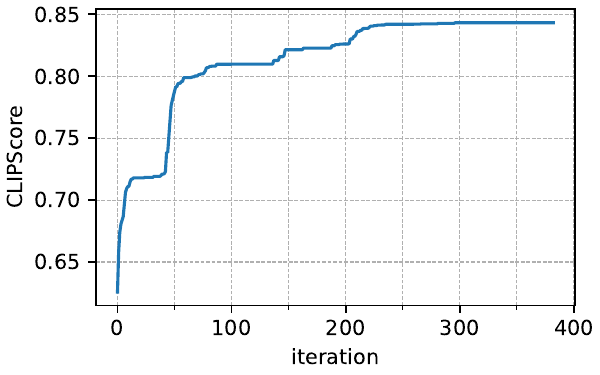}
    \caption{Best \metricClipscore{} in beam for each iteration in beam local search}
    \label{fig:traj:best_score}
\end{figure}

\begin{figure}[t]
    \centering
    \includegraphics[width=\linewidth]{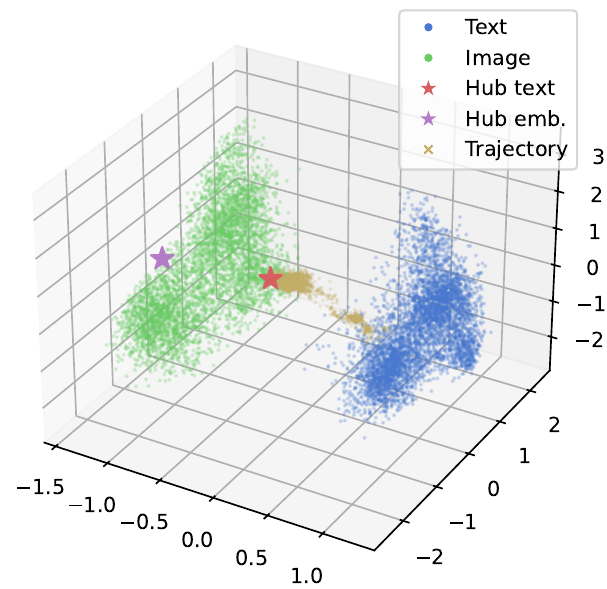}
    \caption{
    Scatter plots of embeddings in MSCOCO
    }
    \label{fig:traj:emb}
\end{figure}

To clarify the behavior of our algorithm, we tracked the trajectory of local search.
\Cref{fig:traj:best_score} shows the best score in the beam for each iteration of our beam local search in MSCOCO.
Because our approach is based on a hill-climbing algorithm, the scores monotonically increased.
In early iterations, the score significantly increased, and then the improvement margin became smaller and converged.

Furthermore, we also investigated the trajectory of local search in the embedding space.
\Cref{fig:traj:emb} plots the embeddings of the test cases, optimal hub, hub text finally obtained with our method, and the trajectory in MSCOCO.
We used principal component analysis (PCA) to reduce the dimensionality of these embeddings to three dimensions for visualization.
As shown here, the hub text left the cluster of text embeddings through local search and moved toward the center of the image cluster.
The finally obtained hub text behaves more like an image within the embedding space, even though the input representation is text, which suggests that this results in unreasonably high similarities to images.
\modify{
From these analyses, we found that the hubness problem may be related to the modality gap problem~\citep{liang-etal-2022-mind,an-etal-2025-i0t}.
Investigating the root cause of the hubness problem and developing methods to mitigate it by leveraging the properties of the embedding space remain important directions for future work.
}

\subsection{Beam size in local search}

\begin{figure}
    \centering
    \includegraphics[width=1.0\linewidth]{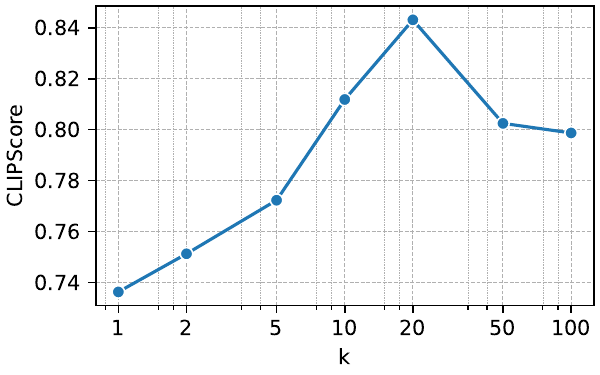}
    \caption{\metricClipscore{} when varying beam size $k$ in MSCOCO validation set}
    \label{fig:beam_size}
\end{figure}

We varied beam sizes $k \in \{ 1, 5, 10, 20, 50, 100 \}$ in the local search, 
with the results shown in \Cref{fig:beam_size}.
For $k \leq 20$, the score increases as the beam size is increased.
While the score decreases for $k \geq 50$ compared to $k = 20$, it is always better than $k=1$.
This demonstrates that our beam local search is more effective than the conventional greedy local search.
In addition, as $k$ increases, computation time grows linearly, and for $k \geq 50$, it became extremely time-consuming.
Therefore, we tuned the beam size $k$ from $\{ 5, 10, 20 \}$ for each model.

\subsection{Complexity and implementation details}
\label{sec:complexity}

\paragraph{Time complexity of local search}
Hub text identification is an NP-hard problem because it is determined by computing the scores of all possible texts in $\spaceText$.
As with the conventional method, the local search is time-consuming.
The time complexity of the beam local search is $\mathcal{O}(\sizeBeam{}T|\vocab||\setDataImage|)$, where $T \in \N$ is the number of iterations until convergence in the main loop.
Since it updates until all tokens are replaced for all beams by checking them with $\bm{\mathcal{P}}$, the number of iterations $T$ roughly scales linearly with the average sentence length.
Our method is slower than the conventional method, due to the addition of a for-loop for the beam search.

\paragraph{Efficient local search using boss--worker pattern}
To leverage the parallel computing capabilities of multiple GPUs, we implemented the beam local search based on the boss--worker design pattern with a single main process and multiple child processes.
The boss submits tasks to a task queue, and each worker associated with the corresponding GPU receives and executes them.
Specifically, a task consists of the current state and a query for the token position that will be replaced next.
Each worker receives a task and computes the evaluation scores of candidate texts where the $i$-th token is replaced with the candidate token.
In the loop of \cref{alg:lbs:vocabloop} in \cref{alg:lbs}, instead of evaluating all candidate texts over the entire vocabulary, each worker is responsible for only a subset of the vocabulary.
The boss aggregates the evaluation scores computed by workers and then updates the beams with the $\funcTopk$ function in \cref{alg:lbs:topk}.
In addition, to reduce the amount of data in the inter-process communication, we implemented a mechanism where each worker returns only the top-$k$ results, i.e., the boss receives only the $k \times \text{\#workers} \ll |\vocab|$ candidates for each beam.

\section{Conclusion}
\label{sec:conclusion}

This study attempts to find the hub embedding and hub text in cross-modal encoders to identify the vulnerabilities of their models.
To this end, we propose methods for analytically finding the optimal hub embedding and finding the hub text more effectively than the prior work, i.e., the beam local search algorithm.
Our experiments with image captioning and image-to-text retrieval tasks showed that the hub text found with the proposed method consistently received higher similarity even with unrelated images compared to both the reference captions and the hub text found with the previous method.
A key finding is that even powerful models exhibiting high correlations with human assessments are susceptible to hub texts.
To develop more reliable models, it is crucial to evaluate not only benchmark scores but also robustness against such malicious attacks.

\section*{Limitations}
\paragraph{Time complexity}
The time complexity of beam local search is high, as discussed in \cref{sec:complexity}.
When using larger encoder models or increasing the beam size, it is extremely time-consuming, even if employing our efficient implementation.

\paragraph{Optimization for sequence length in local search}
The beam local search optimizes only the fixed-length token sequence and does not alter the sequence length.
While searching over variable-length sequences could potentially yield hub texts with higher scores, \modify{
allowing insertion and deletion significantly increases the search space, leading to exponential growth in computational complexity and making it infeasible to complete the search in a realistic time.
Thus, we restrict our algorithm to the fixed-length space of the decoded hub text.
Developing methods for efficient variable-length search remains a challenge for future work.
}

\paragraph{Detectability}
\modify{
The identified hub text is non-natural, so it could be detected by using other encoders or language models.
Our research argues for the necessity of such filtering.
While our results make it seem self-evident that such filtering can prevent these issues, we can consider such filtering because our work has revealed the existence of the hub text across various CLIP models.
This is supported by the fact that real-world search engines, recommendation systems, and question-answering systems do not always incorporate such mechanisms, which further emphasizes the need for this research to advocate for such filtering.
}

\modify{
Furthermore, calculating PPL for all instances is expensive, and over-filtering might degrade performance for the core objective of ``finding relevant instances.'' For example, instances with high PPL due to many technical terms or neologisms may be filtered out, thereby diminishing the usefulness of search engine or recommendation system.
}

\modify{
Developing methods to efficiently filter hub texts while maintaining downstream task performance remains an important challenge for future work.
}

\section*{Ethical Considerations}
\paragraph{Potential for abuse}

While our algorithm could potentially be exploited to generate hub texts, the risk of misuse can be reduced through the use of trusted data sources and responsible system design.
In practice, recent efforts have been made to improve the security of data pipelines.
For example, the \verb|trust_remote_code| option in HuggingFace datasets~\citep{lhoest-etal-2021-datasets}, which previously allowed the execution of arbitrary remote code, is no longer supported as of version 4.0.0.
This change reduces the risk of injecting malicious or contaminated examples into datasets during download.

Through this work, we provide analyses of vulnerabilities caused by the well-known hubness problem, 
with the goal of facilitating research on safeguards and mitigation strategies.
Since hubness has been observed in a wide range of embedding spaces, we report our findings to raise awareness of its potential implications.
Importantly, this study is limited to identifying hub texts that lead to high similarity scores, i.e., it does NOT involve the extraction of personal information or the generation of harmful or malicious content.

We also emphasize the importance of evaluating not only benchmark scores but also safety and robustness against attacks, such as the hub text.
We believe that this study will help identify vulnerabilities of cross-modal encoders and contribute to elucidating the causes and nature of hubness.

\paragraph{Co-ordinated disclosure}
We focused on the vulnerability of cross-modal encoders, falling under the category of coordinated disclosure.
While we successfully identified the hub texts in cross-modal encoders, hubness is already widely known as a weakness and vulnerability in embedding spaces~\citep{radovanovic-etal-2010-hubs}, and several methods for mitigating such weakness have been proposed~\citep{dinu-2015-etal-improving, lazaridou-etal-2015-hubness,huang-etal-2019-hubless,wang-etal-2023-balance, chowdhury-etal-2024-nearest}.
This fact is further supported by a publicly available paper~\citep{deguchi-etal-2026-hacking}, which alarms the vulnerability caused by the hubness problem.
This paper primarily focused on analyzing the well-known hubness problem in detail, including how hubs appear in the concrete text space and are projected into the embedding space.
Thus, our work is not concerned with discovering any new unknown vulnerabilities but rather with investigating the nature of an existing weakness to better understand its mechanisms, aiming to solve the problem at its root.
As such, this paper does not fall under the definition of coordinated disclosure as stated in ``\textit{e. Works detailing new (i.e. previously not public) security weaknesses/failures without any documentation of co-ordinated disclosure may be in breach of policy.}'' of the ACL Policy on Publication Ethics\footnote{
\url{https://www.aclweb.org/adminwiki/index.php/ACL_Policy_on_Publication_Ethics\#Co-ordinated_disclosure}
}.

\bibliography{custom,anthology-1,anthology-2}

\appendix

\section{Licenses}
\label{sec:license}
\paragraph{Models}
\begin{table*}[t]
    \centering
    \small
    \begin{tabular}{@{}lll@{}}
        \toprule
        Model & License & Reference \\
        \midrule
        \texttt{openai/clip-vit-base-patch32} & MIT & \citep{radford-etal-2021-learning} \\
\texttt{openai/clip-vit-large-patch14} & MIT & \citep{radford-etal-2021-learning}\\
\texttt{openai/clip-vit-large-patch14-336} & MIT & \citep{radford-etal-2021-learning}\\
\texttt{laion/CLIP-ViT-L-14-laion2B-s32B-b82K} & MIT & \citep{schuhmann-etal-2022-laionb}\\
\texttt{laion/CLIP-ViT-H-14-laion2B-s32B-b79K} & MIT & \citep{schuhmann-etal-2022-laionb}\\
\texttt{laion/CLIP-ViT-g-14-laion2B-s12B-b42K} & MIT & \citep{schuhmann-etal-2022-laionb}\\
\texttt{apple/DFN2B-CLIP-ViT-L-14} & Apple Machine Learning Research Model & \citep{fang-etal-2024-data}\\
\texttt{apple/DFN5B-CLIP-ViT-H-14} & Apple Machine Learning Research Model & \citep{fang-etal-2024-data}\\
\texttt{apple/DFN5B-CLIP-ViT-H-14-378} & Apple Machine Learning Research Model & \citep{fang-etal-2024-data}\\
\texttt{BAAI/AltCLIP} & CreativeML Open RAIL-M & \citep{chen-etal-2023-altclip} \\
        \texttt{google/mt5-base} & Apache 2.0 & \citep{xue-etal-2021-mt5} \\
        \texttt{Salesforce/blip2-flan-t5-xl} & MIT & \citet{li-etal-2023-blip-2} \\
        \bottomrule
    \end{tabular}
    \caption{Licenses of models we used}
    \label{tab:licenses:models}
\end{table*}

\Cref{tab:licenses:models} lists the licenses and references of the models we employed in our experiments.

\paragraph{Datasets}
In the MSCOCO dataset~\citep{lin-etal-2014-microsoft,karpathy-and-feifei-2015-deep}, the annotations belong to the COCO Consortium and are licensed under a Creative Commons Attribution 4.0 license.
The COCO Consortium does not own the copyright of the images, and use of the images must abide by the Flickr Terms of Use.
The nocaps dataset~\citep{agrawal-etal-2019-nocaps} is released under a CC-BY-2.0 License.
Flickr30k~\citep{young-etal-2014-image} is solely provided for non-commercial research and/or educational purposes.

\section{\modify{Intermediate Texts in Local Search}}
\begin{table*}[t]
    \centering
    \small
    \modifyTable
    \begin{tabular}{@{} l @{}}
    \toprule
    a color photo detfrom the 2 0 1 0 s of kingfish 's quash backpack ' \\
a color photo vc from the 2 0 1 0 s evident kingfish 's quash backpack ' \\
a color photo vc from the 2 0 1 0 s evident kingiller 's quash backpack ' \\
$\vdots$ \\
today color photo \_\_: dishstaged mms bays ], croc xiii \ding{76} trot knogely bw 7 boarded ️: garethapproached cision \\
today color photo \_\_: dishstaged mms middle ], croc air \ding{76} trot maker gely bw 7 boarded ️: garethapproached cision \\
today color photo \_\_: dishstaged mms middle ], croc ée \ding{76} trot maker gely bw 8 boarded ️: garethapproached cision \\
\bottomrule
    \end{tabular}
    \caption{
    \modify{
    Intermediate hub texts of \texttt{openai/clip-vit-base-patch32} in beam local search
    }
    }
    \label{tab:case:ls}
\end{table*}

\modify{
\Cref{tab:case:ls} shows the trajectory of the intermediate hub texts in beam local search.
We only extracted the top-1 candidates for each iteration.
Due to space limitations, we included results only for the first three and the last three steps.
}

\section{Derivation of Optimal Hub Embedding}
\label{sec:derive}

\subsection{Derivation}
\paragraph{Maximize cosine similarity}
Most cross-modal encoders employ cosine similarity for the scoring function, and the optimal hub embedding can be calculated analytically when cosine similarity is used for the scoring function $\funcClipScore$.
For simplicity, we derive it using pure cosine similarity without any scaling and clipping.
From the definition, the optimal hub embedding $\embedding{}^\star \in \R^\sizeDim$ and the objective $\defFunc{\funcObj}{\R^\sizeDim \times 2^{\R^\sizeDim}}{[-1, 1]}$ can be formulated as follows:
\begin{align}
    \embedding{}^\star &\coloneqq \argmax_{\embedding{} \in \R^\sizeDim} \funcObj(\embedding{}; \setDataImage), \\
    \funcObj(\embedding{}; \setDataImage) &\coloneqq \frac{1}{|\setDataImage|} \sum_{\inputImage \in \setDataImage} \frac{\embedding{}^\top \embedding{\inputImage}}{\lVert \embedding{} \rVert \lVert \embedding{\inputImage} \rVert} \\
    &= \frac{\embedding{}^\top}{\lVert \embedding{} \rVert} \cdot \frac{1}{|\setDataImage|} \sum_{\inputImage \in \setDataImage} \frac{\embedding{\inputImage}}{\lVert \embedding{\inputImage} \rVert} \\
    &= \frac{\embedding{}^\top}{\lVert \embedding{} \rVert} \bar{\embedding{}}, \label{eq:derive:ip}
\end{align}
where $\bar{\embedding{}} \coloneqq \frac{1}{|\setDataImage|} \sum_{\inputImage \in \setDataImage} \frac{\embedding{\inputImage}}{\lVert \embedding{\inputImage} \rVert}$, i.e., the averaged vector of all $L^2$-normalized image embeddings in the tuning data $\setDataImage$.
Since \cref{eq:derive:ip} calculates the inner product of two vectors, the upper bound of the absolute value of the inner product can be obtained using Cauchy--Schwarz inequality:
\begin{equation}
    \label{eq:derive:ineq}
    \left| \frac{\embedding{}^\top}{\lVert \embedding{} \rVert} \bar{\embedding{}} \right| \leq \left\lVert \frac{\embedding{}}{\lVert \embedding{} \rVert} \right\rVert \cdot \lVert \bar{\embedding{}} \rVert = \lVert \bar{\embedding{}} \rVert.
\end{equation}
In \cref{eq:derive:ineq}, the equality condition is when $\embedding{}$ and $\bar{\embedding{}}$ are parallel.
Hence, when $\embedding{} \propto \bar{\embedding{}} = \frac{1}{|\setDataImage|} \sum_{\inputImage \in \setDataImage} \frac{\embedding{\inputImage}}{\lVert \embedding{\inputImage} \rVert}$ is satisfied, $\frac{\embedding{}^\top}{\lVert \embedding{} \rVert} \bar{\embedding{}} = \funcObj(\embedding{}; \setDataImage)$ is maximized.

\paragraph{Maximize inner product}
Similar to cosine similarity, the optimal hub embedding in the inner product can also be analytically obtained from image embeddings.
In this case, the objective function can be formulated as follows:
\begin{align}
    \funcObj(\embedding{}; \setDataImage) &\coloneqq \frac{1}{|\setDataImage|} \sum_{\inputImage \in \setDataImage} \embedding{}^\top \embedding{\inputImage} \\
    &= \embedding{}^\top \frac{1}{|\setDataImage|} \sum_{\inputImage \in \setDataImage} \embedding{\inputImage} \\
    &= \embedding{}^\top \bar{\embedding{}}.
\end{align}
The upper bound of the absolute value of the inner product can be obtained, as well as \cref{eq:derive:ineq}:
\begin{equation}
    \left| \embedding{}^\top \bar{\embedding{}} \right| \leq \left\lVert \embedding{} \right\rVert \cdot \lVert \bar{\embedding{}} \rVert = \left\lVert \embedding{} \right\rVert \cdot \lVert \bar{\embedding{}} \rVert \cdot \cos 0.
\end{equation}
Hence, when $\embedding{}$ and $\bar{\embedding{}}$ have the same angle, the objective function will be maximized with a large $L^2$-norm of $\embedding{}$.

\paragraph{Minimize squared Euclidean distance}
In $k$-nearest neighbor search, the squared Euclidean distance is often used\footnote{
For example, Faiss~\citep{johnson-etal-2019-billion,douze-etal-2024-faiss} and Rii~\citep{matsui-etal-2018-reconfigurable} use the squared Euclidean distance for the default distance function.
}.
Since we maximize the objective in this paper, it is defined as follows:
\begin{align}
    \funcObj(\embedding{}; \setDataImage) \coloneqq -\frac{1}{|\setDataImage|} \sum_{\inputImage \in \setDataImage} \lVert \embedding{} - \embedding{\inputImage} \rVert^2.
    \label{eq:derive:msed:obj}
\end{align}
Since $\lVert \embedding{} - \embedding{\inputImage} \rVert^2 = \lVert \embedding{}\rVert^2 - 2 \embedding{}^\top \embedding{\inputImage} + \lVert \embedding{\inputImage} \rVert^2$ and $\frac{1}{|\setDataImage|} \sum_{\inputImage \in \setDataImage}\lVert \embedding{\inputImage} \rVert^2$ is a constant under the objective function, \cref{eq:derive:msed:obj} can be formulated as follows:
\begin{align}
(\ref{eq:derive:msed:obj})
    &\propto -\lVert \embedding{} \rVert^2 +  \frac{1}{|\setDataImage|} \sum_{\inputImage \in \setDataImage} 2 \embedding{}^\top \embedding{\inputImage} \\
    &= -\lVert \embedding{} \rVert^2 + 2 \embedding{}^\top \frac{1}{|\setDataImage|} \sum_{\inputImage \in \setDataImage}  \embedding{\inputImage} \\
    &= -\lVert \embedding{} \rVert^2 + 2 \embedding{}^\top \bar{\embedding{}}.
\end{align}
Since $g\colon \embedding{} \mapsto -\lVert \embedding{} \rVert^2 + 2 \embedding{}^\top \bar{\embedding{}}$ is a convex quadratic function; thus, the global maxima of $g$ can be obtained as follows:
\begin{align}
    \nabla_{\embedding{}} g(\embedding{}) &= \nabla_{\embedding{}} (-\lVert \embedding{} \rVert^2 + 2 \embedding{}^\top \bar{\embedding{}}) \\
    &= -2 \embedding{} + 2 \bar{\embedding{}} = 0, \\
    \therefore \embedding{} &= \bar{\embedding{}}.
\end{align}
Therefore, the optimal hub embedding is the average vector of image embeddings.

\subsection{\modify{Optimality gap between \metricClipscore{} and cosine similarity}}
\modify{
For all experiments, we used the optimal solution in cosine similarity for hub acquisition because we used the same hub text for not only the image captioning evaluation but also the image-to-text retrieval task, which retrieves the $k$-nearest neighbor texts based on cosine similarity.
We discuss the optimality gap between \metricClipscore{} and cosine similarity in this section.
}

\modify{
The differences between the two are scaling and clipping, as shown in \cref{eq:clipscore}.
Since the scaling parameter $\constScale$ is constant with respect to the variable being maximized, the optimal solution is preserved.
In contrast, non-linear zero clipping yields differences in the exact optimal solutions.
The clipping in \metricClipscore{} ignores images that have negative cosine similarity scores with a hub embedding, so the \metricClipscore{} is calculated with images that are contained in the hemisphere centered at a fixed embedding in the hypersphere.
Here, the solution we obtained has positive cosine similarity with all images and is the optimal solution on the hemisphere that contains them.
Thus, we obtained the global optima on the fixed hemisphere that contains all images.
Using other hemispheres means that the \metricClipscore{} for some images will be 0, and the optimized embedding under such hemispheres would not be the hub embedding.
}

\modify{
Moreover, considering all hemispheres, including ones that allow $\metricClipscore{} = 0$ for some images, can be regarded as a central hyperplane arrangement problem~\citep{orlik-terao-1992-arrangements}, and we would need to compute
$2\sum_{i=0}^{\sizeDim{}-1} \binom{|\setDataImage{}|-1}{i}$ times~\citep{zaslavsky-1975-facing}, which is infeasible.
}

\modify{From the above, we obtained the optimal hub embedding based on cosine similarity that works generically across multiple tasks.}

\section{
\modify{
Hub Text Identification Across Other Encoder Types
}
}

\begin{table}[t]
    \centering
    \small
    \modifyTable
    \begin{tabular}{@{} l rr @{}}
    \toprule
    Caption or text & MSCOCO & nocaps \\
    \midrule
    BLIP-2 & 0.354 & 0.352 \\
    Human & 0.381 & 0.380 \\
    GLS & 0.356 & 0.335 \\
    Ours & \textbf{0.406} & \textbf{0.400} \\
    \bottomrule
    \end{tabular}
    \caption{
    \modify{
        Evaluation of image captioning tasks using \texttt{google/siglip2-base-patch16-256}
    }
    }
    \label{tab:clips:corpus_scores:siglip}
\end{table}

\begin{table}[t]
    \centering
    \small
    \modifyTable
    \tabcolsep 2pt
    \begin{NiceTabular}{@{}l rr rr rr rr rr@{}}
    \toprule
    & \multicolumn{2}{@{}c@{}}{NDCG} & \multicolumn{2}{@{}c@{}}{MAP} & \multicolumn{2}{@{}c@{}}{Recall} & \multicolumn{2}{@{}c@{}}{Precision} & \multicolumn{2}{@{}c@{}}{MRR} \\
    \cmidrule(lr){2-3} \cmidrule(lr){4-5} \cmidrule(lr){6-7} \cmidrule(lr){8-9} \cmidrule(l){10-11}
    \#CT & ${}^\text{@}$1 & ${}^\text{@}$10 & ${}^\text{@}$1 & ${}^\text{@}$10	& ${}^\text{@}$1 & ${}^\text{@}$1k & ${}^\text{@}$1 & ${}^\text{@}$5 & ${}^\text{@}$1 & ${}^\text{@}$10 \\
    \midrule
    0 & 69.4 & 61.5 & 13.9 & 50.1 & 13.9 & 99.8 & 69.4 & 49.5 & 69.4 & 77.6 \\
    1 & 54.2 & 56.7 & 10.8 & 44.4 & 10.8 & 99.8 & 54.2 & 45.8 & 54.2 & 68.7 \\
    1,000 & 54.1 & 37.8 & 10.8 & 29.9 & 10.8 & 37.7 & 54.1 & 30.5 & 54.1 & 57.9 \\
    \bottomrule
    \end{NiceTabular}
    \caption{\modify{Image-to-text retrieval performance of \texttt{google/siglip2-base-patch16-256} on MSCOCO}}
    \label{tab:i2t:results:siglip:mscoco}
\end{table}

\modify{
To investigate the generality of our method, we also identify hub texts in SigLIP~\citep{zhai-etal-2023-sigmoid}, where the models are trained with a sigmoid-based loss function that differs from CLIP.
We used \texttt{google/siglip2-base-patch16-256}~\citep{tschannen-etal-2025-siglip} for the experiments.
\Cref{tab:clips:corpus_scores:siglip} and \cref{tab:i2t:results:siglip:mscoco} show the results of the caption evaluation and image-to-text retrieval tasks, respectively.
These results demonstrated that we also successfully identified the hub text in SigLIP-2 trained using a different loss function.
We confirmed that robustness against hubness still remains an issue, even with the latest model.
}

\section{Comparison of Methods for Hub Text Identification}

\begin{table}[t]
    \centering
    \small
    \tabcolsep 2.5pt
    \begin{tabular}{@{}lll@{}}
    \toprule
         & \citet{deguchi-etal-2026-hacking} & Ours \\
         \midrule
        Target metric & COMET & \metricClipscore{} \\
        Scoring function & Feed-forward network & Cosine similarity \\
        Hub embedding & Numerical solution & Analytical solution \\
        Local search & Greedy search & Beam search \\
        Search order & Sequential & Random \\
        Method type & White-box & Black-box \\
    \bottomrule
    \end{tabular}
    \caption{Comparison of related work~\citep{deguchi-etal-2026-hacking} and our work}
    \label{tab:method:compare}
\end{table}

\Cref{tab:method:compare} lists the differences between the related work~\citep{deguchi-etal-2026-hacking} and our work.


\section{Dataset Statistics}

\begin{table}[t]
    \centering
    \begin{tabular}{@{}lll@{}}
    \toprule
        Dataset & \#instances \\
        \midrule
        MSCOCO (Train) & 113,287 \\
        MSCOCO (Validation) & 5,000 \\
        MSCOCO (Test) & 5,000 \\
        nocaps & 4,500 \\
        Flickr30k & 1,000 \\
        \bottomrule
    \end{tabular}
    \caption{Statistics of datasets we used}
    \label{tab:datastats}
\end{table}
The dataset statistics are listed in \cref{tab:datastats}.

\end{document}